\documentclass[journal]{IEEEtran}
\IEEEoverridecommandlockouts

\usepackage{cite}
\usepackage{amsmath,amssymb,amsfonts}
\usepackage{algorithmic}
\usepackage{graphicx}
\usepackage{textcomp}
\usepackage{xcolor}
\usepackage{tabularx}
\usepackage{diagbox}
\usepackage{siunitx}
\usepackage{adjustbox}
\usepackage{siunitx}
\usepackage{multirow}
\usepackage{booktabs}
\usepackage{array}
\usepackage{multirow}
\usepackage{setspace} 
\usepackage{arydshln}

\begin{document}

\title{Physics-informed Deep Learning for Muscle Force Prediction with Unlabeled sEMG Signals\\

\author{Shuhao Ma, Jie Zhang,~\IEEEmembership{Member, IEEE},
        Chaoyang Shi,~\IEEEmembership{Member, IEEE},
        Pei Di,~\IEEEmembership{Member, IEEE},
        Ian D. Robertson,~\IEEEmembership{Fellow, IEEE},
        Zhi-Qiang Zhang,~\IEEEmembership{Member, IEEE}}

\thanks{This work was supported in part by the UKRI Horizon Europe Guarantee under Grant EP/Y027930/1, in part by the Royal Society under Grant IEC/NSF/211360, in part by the EU Marie Curie Individual Fellowship under Grant 101023097, and in part by the China Scholarship Council (CSC) under Grant 202208320117. (Corresponding author: Zhi-Qiang Zhang)}
\thanks{{\large ${\ast}$} The first two authors contributed equally to this work.}    
\thanks{Shuhao Ma, Ian D. Robertson, and Zhi-Qiang Zhang are with the School of Electronic and Electrical Engineering, University of Leeds, Leeds LS2 9JT, U.K. (e-mail: elsma@leeds.ac.uk; i.d.robertson@leeds.ac.uk; z.zhang3@leeds.ac.uk).}
\thanks{Jie Zhang is with the Centre for Wireless Innovation, Queen's University Belfast, Belfast BT3 9DT, U.K. (e-mail: jie.zhang@qub.ac.uk).}
\thanks{Chaoyang Shi is with the Key Laboratory of Mechanism Theory and Equipment Design of Ministry of Education, School of Mechanical Engineering, Tianjin University, Tianjin 300072, China, and also with the International Institute for Innovative Design and Intelligent Manufacturing of Tianjin University in Zhejiang, Shaoxing 312000, China (e-mail: chaoyang.shi@tju.edu.cn).}
\thanks{Pei Di is with Rex Bionics Ltd, PO Box 316-063
Auckland 0760, New Zealand  (e-mail: pei.di@imaxhealth.com).}
}
\maketitle

\begin{abstract}
Computational biomechanical analysis plays a pivotal role in understanding and improving human movements and physical functions. Although physics-based modeling methods can interpret the dynamic interaction between the neural drive to muscle dynamics and joint kinematics, they suffer from high computational latency. In recent years, data-driven methods have emerged as a promising alternative due to their fast execution speed, but label information is still required during training, which is not easy to acquire in practice. To tackle these issues, this paper presents a novel physics-informed deep learning method to predict muscle forces without any label information during model training. In addition, the proposed method could also identify personalized muscle-tendon parameters. To achieve this, the Hill muscle model-based forward dynamics is embedded into the deep neural network as the additional loss to further regulate the behavior of the deep neural network. Experimental validations on the wrist joint from six healthy subjects are performed, and a fully connected neural network (FNN) is selected to implement the proposed method. The predicted results of muscle forces show comparable or even lower root mean square error (RMSE) and higher coefficient of determination compared with baseline methods, which have to use the labeled surface electromyography (sEMG) signals, and it can also identify muscle-tendon parameters accurately, demonstrating the effectiveness of the proposed physics-informed deep learning method.
\end{abstract}


\IEEEpeerreviewmaketitle

\section{Introduction}\label{intro}

Human movements need the coordinated actions of various muscle elements, thus accurate muscle force estimation could support promising applications in diverse domains, ranging from efficacious rehabilitation protocol design \cite{1605264}, optimizing motion control \cite{5723418, RN95}, to enhancing clinical decision-making \cite{8371283, KARTHICK201845, 8637973} and the performance of athletes \cite{9546647, app11041450}. The majority of muscle force estimation methods are based on physics-based modeling techniques. For instance, inverse dynamics techniques have been validated to generate reasonable estimations of muscle forces and muscular activation patterns usually based on static optimization \cite{ZARGHAM2019223, RN87, RN89, AMBROSIO20114, RN91}. The static optimization could find the set of muscle forces by minimizing the physiological criterion, such as muscle activation, volume-scaled activation, forces, stresses, metabolic energy or joint contact forces. However, it is challenging to provide the biologically consistent rationale for the selection of any objective function \cite{https://doi.org/10.1002/jor.20876, MODENESE20112185}, due to the lack of knowledge about the method used by the central nervous system \cite{RN96}. Furthermore, physics-based modeling methods also suffer from high computational latency, especially in complex modeling scenarios \cite{TRINLER201955, 4912337}.

To address the time-consuming issue of physics-based methods, data-driven methods have been investigated to establish relationships between the movement variables and neuromuscular status, such as from electromyography (EMG) signals to muscle forces, in the past few years \cite{9380441, 10288253, 10.1371/journal.pone.0247883, 9252155}. Although the training of deep neural networks may be lengthy, as the inference only involves a relatively simple forward pass through the network, it is computationally inexpensive and thus very quick.
For instance, Hua et al. \cite{RN107} proposed a linear regression (LR) and long short-term memory (LSTM)-integrated method (LR-LSTM) to predict the muscle force under the isometric contraction state. Tang et al. \cite{9606872} developed a modified framework to accurately predict muscle forces based on encoder-decoder networks.
Moreover, Lu et al. \cite{Lu_2021} designed an integrated deep learning framework that combined a convolutional neural network (CNN) and a bidirectional LSTM (BiLSTM), complemented by an attention mechanism, for elbow flexion force estimation.
However, all these models are established without explicit physical modeling of the underlying neuromechanical processes, and these conventional “black-box” tools do not consider the physical significance underlying the modeling process \cite{9736635, RN97}.

In recent years, the integration of physics-based modeling and data-driven modeling has emerged as an effective strategy to overcome the limitations of these two methods, such as deep energy method-based deep neural network \cite{SAMANIEGO2020112790}, deep Ritz method \cite{RN116}, physics-informed deep neural operator networks \cite{doi:10.1126/sciadv.abi8605}, and thermodynamics-informed neural network \cite{HERNANDEZ2023115912}, etc. 
In musculoskeletal (MSK) modeling, some existing works also investigate the integration of physics domain knowledge and data-driven modeling. Specifically, Zhang et al. \cite{9970372} proposed a physics-informed deep learning framework for muscle forces and joint kinematics prediction, in which the equation of motion was embedded into the loss function as the soft constraints to penalize and regularize the deep neural network training. They also designed a physics-informed deep transfer learning framework to strengthen the performance of the personalized MSK modeling \cite{9975275}. Taneja et al. \cite{10.1115/1.4055238} designed a novel physics-informed parameter identification neural network for simultaneously predicting motion and identifying parameters of MSK systems. They also developed a multi-resolution physics-informed recurrent neural network to further enhance motion prediction and parameter identification \cite{RN112}. Shi et al. \cite{10375088} developed a physics-informed low-shot learning approach based on generative adversarial network for muscle forces and joint kinematics prediction, which first integrated the Lagrange’s equation of motion into the generative model to restrain the structured decoding of discriminative features, and a physics-informed policy gradient was then proposed to enhance the adversarial learning efficiency by rewarding the consistent physical representation of extrapolated estimations and physical references. Although the aforementioned physics-informed data-driven methods have achieved great progress for MSK modeling enhancement, there are still two main challenging issues: 1) Labeled data are required for model training \cite{9970372, 9975275, 10375088}, 2) For muscle force prediction, \cite{10.1115/1.4055238} and \cite{RN112} need to reprocess the network's output in conjunction with the MSK dynamics, making the running latency far over the maximum 75 ms considered optimal real-time biofeedback. Therefore, it is urgent to design a novel physics-informed neural network framework, that does not need to acquire a large amount and sufficient labeled data for deep neural network training, and can still work well in real-time application scenarios. 

In this paper, a novel physics-informed deep learning method is presented to predict muscle forces using unlabeled surface EMG (sEMG) data. Additionally, the proposed method could also identify muscle-tendon parameters of the Hill muscle model. 
In the proposed method, a fully connected neural network (FNN) is utilized to implement the designed physics-informed deep learning framework, and the Hill muscle model is embedded into FNN as the additional loss component to further penalize and regularize the behavior of FNN.
To validate the proposed method, a self-collected dataset consisting of six healthy subjects performing wrist flexion/extension motion is used in the experiments. According to the experimental results, the proposed method with unlabeled sEMG data shows comparable and even better performance compared with selected machine learning and deep learning methods, which have to use labeled sEMG data. 

\begin{figure*}
  \centering
  \includegraphics[scale = 1]{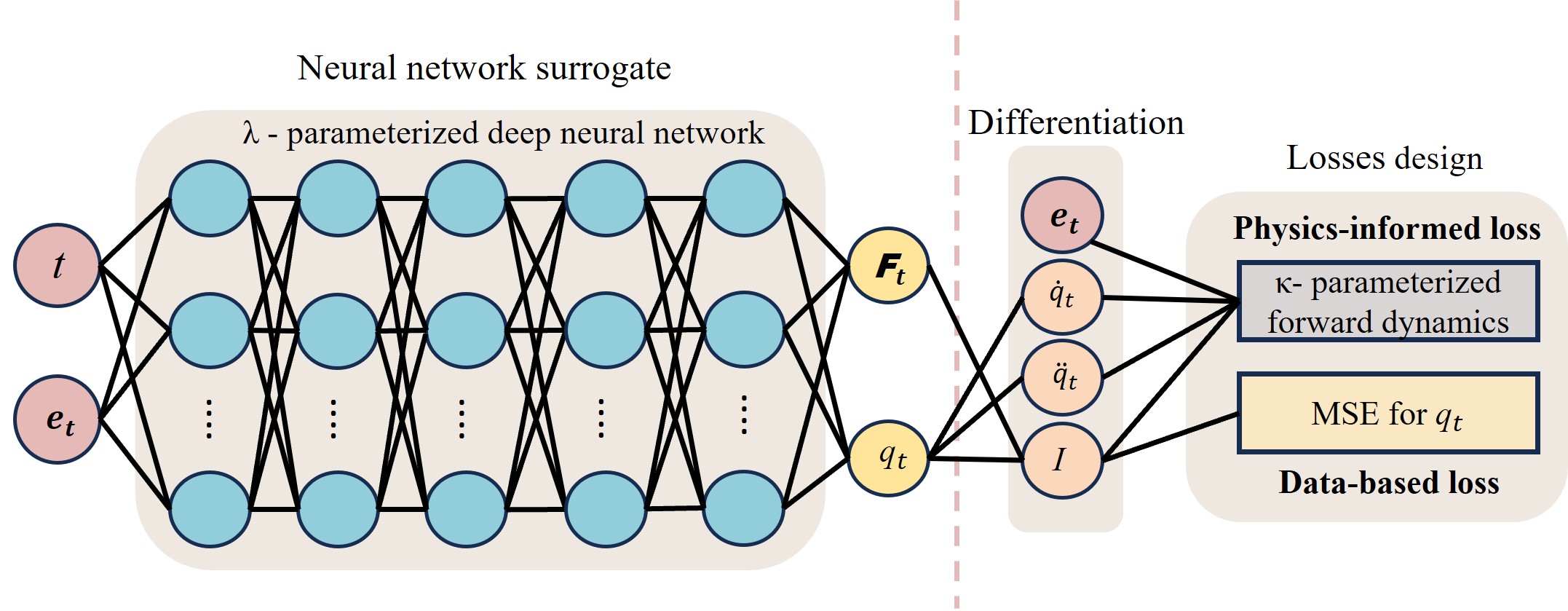} 
  \caption{Main framework of the proposed method. Inputs to the $\lambda$-parameterized deep neural network are sEMG measurements $e_t = (e_{t,1}, e_{t,2},\cdots, e_{t, N})$ and time $t$, while outputs are joint movements $q_t$ and muscle forces $F_t = (F_{t,1}, F_{t,2},\cdots, F_{t, N})$, where $N$ is the total number of muscles at the joint of interest. For the subject-specific Hill-muscle-based forward dynamics model, $\kappa = (A, \kappa_1, \kappa_2,\cdots, \kappa_N)$, where $n = (1,2,\cdots, N)$, $\kappa_n$ is the muscle-tendon parameters of the $n$th muscle, and $A$ is the EMG-to-activation coefficient.}
  \label{fig:0}
\end{figure*}

The remainder of this paper is organized as follows: The proposed physics-informed deep learning method is detailed in Section \ref{method}, including the main framework, the network architecture and training strategy, the loss function, and the incorporation of Hill-muscle-based forward dynamics. Dataset and experimental settings are described in Section \ref{Dataset}. Experimental results are reported in Section \ref{results}, and discussions are presented in Section \ref{discussions}. Finally, conclusions are given in Section \ref{conclusion}.

\section{Methods}\label{method}
In this section, we first describe the details of the proposed method, in the context of muscle force prediction and muscle-tendon parameters identification from 
unlabeled sEMG signals, including the main framework, the network architecture and training, the loss function as well as the incorporation of Hill-muscle-based forward dynamics. 

\subsection{Main Framework} 
Fig. \ref{fig:0} shows the main framework of the proposed method, in the context of muscle forces prediction and muscle-related physiological parameters identification from unlabeled sEMG signals. Specifically, in the neural network surrogate, inputs to the $\lambda$-parameterized deep neural network are sEMG measurements and the corresponding time $t$, while outputs are the joint movement $q_t$ and muscle forces $F_t = (F_{t,1}, F_{t,2},\cdots, F_{t, N})$, where $N$ is the total number of muscles at the joint of interest. A FNN is utilized to extract more discriminative features and build the relationship between the inputs and outputs. Different from conventional loss functions, the novel total loss consists of the data-based loss and physics-informed losses. The data-based loss is based on mean squared error (MSE), while the physics-informed losses are based on the $\kappa$-parameterized underlying Hill-muscle-based forward dynamics, where $\kappa = (A, \kappa_1, \kappa_2, \cdots, \kappa_N)$ and $A$ is the EMG-to-activation coefficient.

\subsection{FNN Architecture and Training}
Without loss of generality, a FNN is utilized as the deep neural network to implement the proposed method, and it is composed of four fully connected (FC) blocks and one regression block. To be specific, each FC block has one linear layer, one ReLU layer and one dropout layer. The regression block consists of one ReLU layer and one dropout layer. The trainable parameters of FNN are obtained by minimizing the loss function (more details about the loss function refer to Section II-C). The training is performed using the Adam algorithm with an initial learning rate of 0.001, the batch size is 1, the maximum iteration is 1000, and the dropout rate is 0.3.

\subsection{Loss Function Design}
The designed loss function of the proposed method includes the data-based loss $L_q$, and physics-informed losses $L_{fd}$ and $L_{F}$, which can be represented as
\begin{equation}
L_{total} = L_q + L_{fd} + L_{F}
\end{equation}
where $L_q$ is the MSE of the actual joint angles and predicted joint angles, 
$L_{fd}$ represents the Hill-muscle-based forward dynamics constraint,
$L_{F}$ is an implicit relationship between muscle forces predicted by the neural network and calculated by the embedded Hill muscle model. 

\subsubsection{MSE Loss} The MSE of ground truths of the joint angle and the joint angle predicted by FNN is
\begin{equation}
L_q = \frac{1}{T} \sum_{t=1}^{T}(\hat{q_t} - q_t)^2
\end{equation}
where $q_t$ is as the ground truth of the joint angle and $\hat{q_t}$ is the predicted joint angle of FNN with the trainable parameters $\lambda$ at time $t$. 

\subsubsection{Physics-informed Forward Dynamics Loss} $L_{fd}$ reflects underlying relationships among the muscle force and kinematics in human motion, which can be written as
\begin{align}
L_{fd} = \frac{1}{T} \sum_{t=1}^{T} (M(\hat{q_t})\ddot{\hat{q_t}} + C(\hat{q_t}, \dot{\hat{q_t}}) + G(\hat{q_t}) - \tau_t(\kappa))^{2}
\label{r1}
\end{align}
where $M(\hat{q_t}), C(\hat{q_t}, \dot{\hat{q_t}})$ and $G(\hat{q_t})$ are the mass matrix, the Centrifugal and Coriolis force, and the gravity. $\dot{\hat{q_t}}$ and $\ddot{\hat{q_t}}$ are the predicted joint angular velocity and joint angular acceleration. $\tau_t(\kappa)$ represents the joint torque, which is calculated by the summation of the product of the moment arm and muscle-tendon force:
\begin{equation}
\tau_t(\kappa) =\sum_{n=1}^{N} F_{t,n}^{mt}(\kappa_n)r_{t,n}.
\label{torque}
\end{equation}
where $N$ is the number of muscles involved, $r_{t,n}$ is the moment arm of the $n$th muscle which can be calculated using the polynomial equation and the scale coefficient against joint angle $q_t$\cite{RAMSAY2009463}, $F_{t,n}^{mt} (\kappa_n)$ is the estimated muscle force by the Hill muscle model with muscle-tendon parameters $\kappa_n$ (Additional details about the calculation of the muscle force $F_{t,n}^{mt} (\kappa_n)$ are located in Section \ref{hillmuscleforce}).

\subsubsection{Physics-informed Implicit Loss} There is also an implicit relationship between the muscle forces $\hat{F}_n^t$ predicted by FNN and the muscle force $F_{t,n}^{mt} (\kappa_n)$ calculated by the Hill muscle model. Thus, $L_{F}$ is designed for estimating muscle forces by minimizing the difference between $\hat{F}_n^t$ and $F_{t,n}^{mt} (\kappa_n)$, which can be written as
\begin{equation}
L_{F}= \frac{1}{T} \sum_{t=1}^{T}\sum_{n=1}^{N}(\hat{F}_n^t - F_{t,n}^{mt}(\kappa_n))^2
\end{equation}

Therefore, the optimal neural network parameters $\lambda$ and the subject-specific physiological parameters $\kappa$ can be obtained by minimizing the composite loss function $L_{total}$:
\begin{equation}
\hat{\kappa}, \hat{\lambda} = \underset{\kappa, \lambda}{\arg \min} (L_{total}).
\label{para}
\end{equation}

\subsection{Hill Muscle Force Estimation}\label{hillmuscleforce}

For the $n$th muscle-tendon unit, its muscle-tendon parameters $\kappa_n$ include the isometric muscle force $F_{o,n}^{m}$, the optimal muscle length ${l}_{o,n}^{m}$, the maximum contraction velocity $v_{o,n}$, the tendon slack length $l^{t}_{s, n}$ and the optimal pennation angle $\varphi_{o,n}$,  $\kappa_n = (F_{o,n}^{m}, {l}_{o,n}^{m}, v_{o,n}, l^{t}_{s, n}, \varphi_{o,n})$, and the EMG-to-activation coefficient $A$. 

The Hill-muscle-based forward dynamics model includes activation dynamics and contraction dynamics. 
Activation dynamics refer to the process of transforming pre-processed  sEMG signals $e_{t, n}$ into muscle activation signals $a_{t, n}$, which can be estimated by \cite{9258965}
\begin{equation}
a_{t, n} = \frac{e^{Ae_{t, n}} - 1}{e^A - 1}
\label{eq}
\end{equation}

Muscle forces will be determined, once muscle activation signals $a_{t, n}$ have been obtained. Contraction dynamics used in this study are described by the rigid musculotendon model \cite{10.1115/1.4023390}, in which the pennated muscle element, comprising a contractile element in parallel with a passive elastic element, is connected to an inextensible tendon element. Therefore, the muscle-tendon force can be calculated \cite{RN40}:
\begin{equation}
\begin{aligned}
F^{mt}_{t,n}(\kappa_n) &= (F_{t,n}^{CE} + F_{t,n}^{PE})\cos{\varphi_{t,n}}\\
&= F_{o,n}^{m}(a_{t,n}f_{v}(\overline{v}_{t,n})f_{a}(\overline{l}^{m}_{t,n}) \\
&+ f_{p}(\overline{l}^{m}_{t,n}))cos{\varphi_{t,n}}
\label{Fmt}
\end{aligned}
\end{equation}

\begin{equation}
\varphi_{t,n} = \sin^{-1}(\frac{l^m_{o,n}\sin{\varphi_{o,n}}}{l^{m}_{t,n}})
\label{fi}
\end{equation} 

\begin{equation}
l^m_{t,n} = (l^{mt}_{t,n} - l^{t}_{t,n}){\cos^{-1}{\varphi_{t,n}}}
\label{lm}
\end{equation}
where $F_{t,n}^{CE}$ and $F_{t,n}^{PE}$ are the active force generated by the muscle contraction and the passive force generated by the muscle stretch, respectively.
The pennation angle $\varphi_{t,n}$ is the angle between the orientation of the muscle fiber and tendon, and the pennation angle at the current muscle fiber length $l^{m}_{t,n}$ can be calculated through Eq. (\ref{fi}).
To update the muscle length $l^m_{t,n}$, the muscle–tendon length $l^{mt}_{t,n}$ is approximated by the higher-order polynomial with respect to the predicted joint angle $q_t$, which is exported from OpenSim \cite{10.1371/journal.pcbi.1006223}. $l^{t}_{t,n}$ is the tendon length, and ${v}_{t,n}$ is the contraction velocity which is defined as the time derivative of muscle fiber length.
$f_{a}(\overline{l}^{m}_{t,n})$, $f_{v}(\overline{v}_{t,n})$ and $f_{p}(\overline{l}^{m}_{t,n})$ interpret the force-length-velocity characteristics relating to $a_{t,n}$ and normalized muscle length $\overline{l}^{m}_{t,n}$. 

Before the model training, all the physiological parameters included in $\kappa$ need to be initialized by linear scaling based on the initial values of the generic model from OpenSim. These parameters will be continuously updated in each iteration during the model training process.


\section{Dataset and Experimental Settings}\label{Dataset}
In this section, data collection and preprocessing are first detailed, physiological parameters used in this study, evaluation criteria and baseline methods are then presented, respectively.

\subsection{Data Collection and Preprocessing}
As approved by the MaPS and Engineering Joint Faculty Research Ethics Committee of the University of Leeds (MEEC18-002), this study involves the participation of six subjects who have all provided signed consent forms. We collected data on the subjects' weight and the length of their hands to calculate the moment of inertia of their hands.

During the data collection process, participants were instructed to maintain a straight torso with their shoulder abducted at a $\ang{90}$ angle and their elbow joints flexed at a $\ang{90}$ angle. The continuous wrist flexion/extension motion was recorded using the VICON motion capture system, which tracked joint angles at a rate of 250 Hz using 16 reflective markers on the upper limb. In the meantime, sEMG signals were recorded by Avanti Sensors at a rate of 2000 Hz from the primary wrist muscles, including the Flexor Carpi Radialis (FCR), Flexor Carpi Ulnaris (FCU), Extensor Carpi Radialis Longus (ECRL), Extensor Carpi Radialis Brevis (ECRB), and Extensor Carpi Ulnaris (ECU). The sEMG signals and motion data were synchronized and resampled at a rate of 1000 Hz. 
Each participant completed two repetitive trials at different speeds with a three-minute break between the speed changes to prevent muscle fatigue \cite{9964269}.

The collected sEMG signals underwent a series of processing steps, which included band-pass filtering (20 Hz to 450 Hz), full-wave rectification, and low-pass filtering (6 Hz). Subsequently, these signals were normalized based on the maximum voluntary contraction recorded prior to the experiments, resulting in enveloped sEMG signals. Each trial involving wrist movement included data on time $t$, sEMG signals, and wrist joint angles. The muscle forces calculated by the computed muscle control (CMC) tool from OpenSim were used as ground truths in the experiments.

\subsection{Initialization of Physiological Parameters}
Among the physiological parameters of the muscle-tendon units involved, we choose the maximum isometric muscle force $F_{0, n}^{m}$ and the optimal muscle fiber length $l_{0, n}^{m}$ for the identification. The nonlinear shape factor $A$ in the activation dynamics also needs to be identified. 
Other physiological parameters are obtained by linear scaling based on the initial values of the generic model from OpenSim. Table \ref{tab:table0} shows the details of the initialization of all the physiological parameters of a specific subject as an example. Since there may be differences in terms of magnitude and scale between each parameter due to their different physiological natures, it is necessary to normalize them before training.
\begin{table}[ht]
\scriptsize
\centering
\caption{Physiological parameters involved in the forward dynamics setup of wrist flexion-extension motion for specific subject}
\resizebox{0.46\textwidth}{!}{
\begin{tabular}{@{}c*{6}{c}@{}}
\toprule[\heavyrulewidth]
Parameters       & FCR   & FCU   & ECRL  & ECRB  & ECU    \\ \midrule
$F_{0, n}^{m}$(N)   & 407   & 479   & 337   & 252   & 192    \\
$l_{0, n}^{m}$(m)   & 0.062 & 0.051 & 0.081 & 0.058 & 0.062  \\
$v_{0, n}$(m/s)       & 0.62  & 0.51  & 0.81  & 0.58  & 0.62   \\
$l_{s, n}^{t}$(m)   & 0.24  & 0.26  & 0.24 & 0.22  & 0.2285 \\
$\varphi_{0, n}$(rad) & 0.05  & 0.2   & 0   & 0.16  & 0.06   \\
\midrule
A                & \multicolumn{5}{c}{0.01}               \\ 
\bottomrule[\heavyrulewidth]
\end{tabular}
}
\label{tab:table0}
\end{table}

\begin{figure*}[ht] 
    \centering 
    \includegraphics[width=0.95\textwidth]{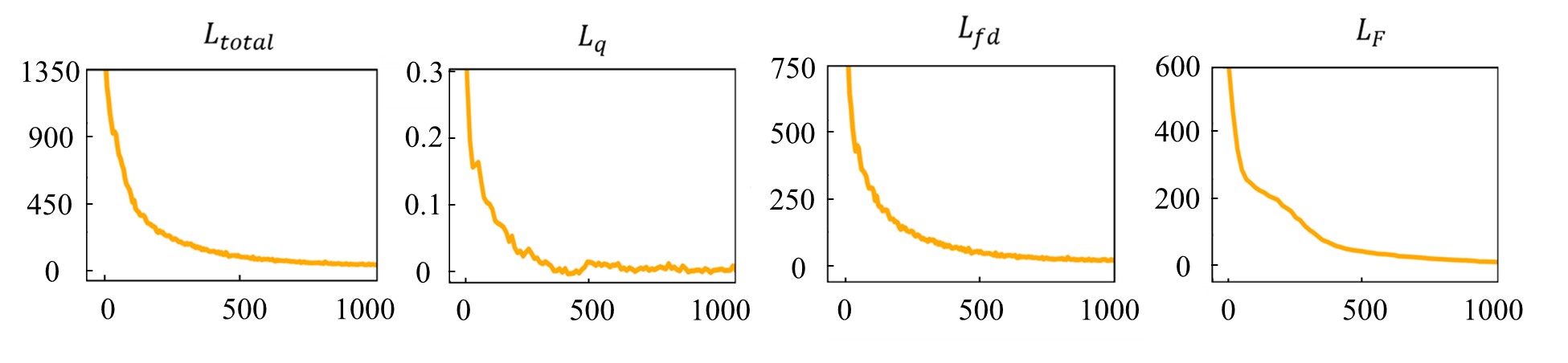}
    \caption{Illustration of different loss terms over the number of iterations.}
    \label{fig:lossconvergence}
\end{figure*}

\subsection{Evaluation Criteria}
In the experiments, root mean square error (RMSE) and coefficient of determination $R^{2}$ are considered as the evaluation criteria to quantify the performance of the proposed method. RMSE is
\begin{equation}
RMSE = \sqrt{{\frac{1}{T}\sum_{t=1}^{T}(U_t - \hat{U}_t})^2} 
\end{equation}
where $T$ is the number of samples, $U_{t}$ and $\hat{U_{t}}$ indicate the ground truth and the predicted value at time $t$, respectively.

$R^{2}$ could be calculated by 
\begin{equation}
\label{r2}
R^2 = 1 - \frac{\sum_{i=1}^{T}(U_t - \hat{U}_t)^2}{\sum_{t=1}^{T}(U_t - \overline{U}_t)^2}
\end{equation}
where $\overline{U}_t$ denotes the mean value of all the samples.

\subsection{Baseline Methods}
To verify the effectiveness of the proposed method, we select LSTM, gated recurrent unit (GRU), CNN, FNN, support vector regression (SVR) and extreme learning machine (ELM) as baseline methods in the experiments. Specifically, the hidden dimensional of LSTM and GRU is 64, and the number of layers is 2, and the batch size of them is 8. CNN has convolutional layers and one FC layer. For each convolutional layer, the kernel size, stride, and padding number are 3, 1 and 3, respectively. The Adam optimizer is employed for CNN training, the batch size is set as 8. FNN has four FC blocks and two regression blocks but without the physics-informed component. Adam optimizer is employed for FNN training, the batch size is set as 1, and the maximum iteration is set as 1000. The radial basis function (RBF) is selected as the kernel function of SVR, and the parameter $C$, which controls the tolerance of the training samples, is set as 100, and the kernel function parameters $\gamma$, which controls the range of the kernel function influence, is set as 1. ELM is a kind of single hidden layer feed-forward neural network with randomly generated hidden layer parameters, its hidden node number is 64 and the Sigmoid function is utilized as the activation function.

\section{Results}\label{results}

In this section, we evaluate the performance of the proposed method using the self-collected dataset. 
The convergence of loss terms is first illustrated, and the parameter identification is then demonstrated. Next, the overall comparisons depict the outcomes of both the proposed method and baseline methods. The robustness and generalization of the proposed method are also investigated, including the performance in the intrasession scenario, effects of network architectures and parameters, and training data number.
The proposed method and all the baseline methods are carried out under the framework of PyTorch, they are implemented on a laptop with a GeForce RTX 3070 Ti graphics card and 32 GB RAM. 

\subsection{Demonstration of Loss Function Convergence}
Fig. \ref{fig:lossconvergence} shows the convergence of different loss terms. 
According to Fig. \ref{fig:lossconvergence}, we can observe that despite the differences in the final convergence values, these four loss terms demonstrate remarkably consistent convergence trends throughout the entirety of the training process. Specifically, all these loss terms could converge after about 800 iterations and finally converge with fast speeds, indicating the effectiveness of the proposed loss function.
Furthermore, the total loss $L_{total}$, as well as $L_{fd}$ and $L_F$, exhibit a smooth and stable convergence pattern throughout the training period. 
In contrast, the MSE loss $L_q$ shows rapid convergence within the initial 200 epochs, followed by slight oscillations. This oscillation could be attributed to the relatively small absolute magnitude of $L_q$.

\subsection{Evaluation of Physiological Parameter Identification}

The subject-specific physiological parameters are identified during the training of the proposed method. Table \ref{tab:table1} presents the estimation and physiological range of the parameters of a specific subject as an example. 
Physiological ranges of the parameters are chosen according to \cite{RN86}. The ranges of the maximum isometric force $F^{m}_{0}$ are set as $\pm \SI{50}{\%}$ of the initial guess, while the ranges of the optimal muscle fiber length $l^{m}_{0}$ are set as $\pm 0.01$ of the initial guess (Details of the initial guesses of these physiological parameters refer to Table \ref{tab:table0}). The identified physiological parameters by the proposed method are all within the physiological range and possess physiological consistency. 
The identified muscle activation dynamics parameter A is -2.29, which is physiologically acceptable in the range of -3 to 0.01.

\begin{figure*}
    \centering
    \includegraphics[width=\textwidth]{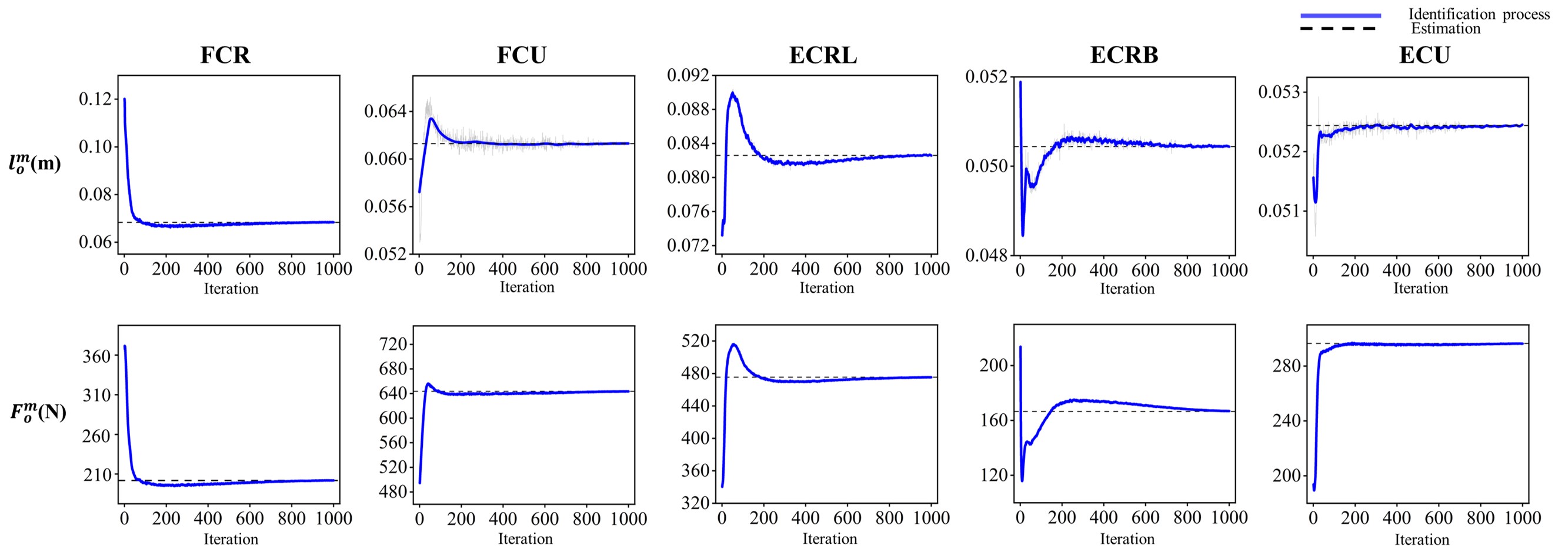}
    \caption{Evolution of the maximum isometric muscle force $F_0^m$ and the optimal muscle fiber length $l_0^m$ identified of the specific subject during the training of the proposed method. The estimations are all within the physiological range and possess physiological consistency.}
    \label{fig:2}
\end{figure*}

\begin{table}[ht]
\LARGE 
\centering
\renewcommand{\arraystretch}{1.2}  
\caption{Identified physiological parameters of the specific subject (Wrist case)}
\resizebox{0.5\textwidth}{!}{
\begin{tabular}{@{}c*{6}{c}@{}}
\toprule
\multirow{1}{*}{Parameter Indices} & \multirow{1}{*}{Muscle Indices} & \multirow{1}{*}{Estimations} & \multirow{1}{*}{Physiological Ranges} \\
\hline
\multirow{5}{*}{${l}_{0}^{m}$(m)} & FCR          & 0.056      & 0.052 - 0.072       \\
                   & FCU          & 0.061      & 0.041 - 0.061       \\
                   & ECRL         & 0.082      & 0.071 - 0.091       \\
                   & ECRB         & 0.050      & 0.048 - 0.068       \\
                   & ECU          & 0.052      & 0.052 - 0.072       \\ \hline
\multirow{5}{*}{${F}_{0}^{m}$(N)} & FCR          & 205.2      & 203.5 - 610.5       \\
                   & FCU          & 644.1      & 239.5 - 718.5       \\
                   & ECRL         & 475.2      & 168.5 - 505.5       \\
                   & ECRB         & 166.7      & 126 - 378           \\
                   & ECU          & 286.6      & 96 - 288            \\ 
\bottomrule
\end{tabular}
}
\label{tab:table1}
\end{table}

\begin{table*}[ht] 
\label{tab:overallrmse}
\centering
\caption{RMSE of the proposed method and baseline methods of muscle forces prediction (wrist case)}
\resizebox{\textwidth}{!}{
\begin{tabular}{c|cccccc|c|ccccccc}
\hline
Subjects             & Methods                   & FCR     & FCU     & ECRL    & ECRB    & ECU      & Subjects             & Methods                   & FCR     & FCU     & ECRL    & ECRB   & ECU     \\ \hline
\multirow{7}{*}{S1} & \multicolumn{1}{c|}{Ours} & 4.99  & 4.62  & 4.35  & 5.33  & 3.12    & \multirow{7}{*}{S4} & \multicolumn{1}{c|}{Ours} & 5.54  & 5.21  & 5.31  & 3.58 & 3.92    \\
 & \multicolumn{1}{c|}{LSTM} & 4.13  & 2.73  & 3.65  & 4.13  & 3.37   &                     & \multicolumn{1}{c|}{LSTM} & 4.32  & 3.69  & 4.97  & 4.71 & 5.02    \\
                    & \multicolumn{1}{c|}{GRU}  & 4.27  & 3.54  & 2.99  & 4.86  & 4.52   &                     & \multicolumn{1}{c|}{GRU}  & 5.63  & 4.11  & 6.16  & 4.97 & 4.37   \\
                    & \multicolumn{1}{c|}{CNN}  & 4.61  & 4.11  & 3.37 & 5.19  & 3.16   &                     & \multicolumn{1}{c|}{CNN}  & 6.62  & 5.37  & 5.29  & 5.01 & 6.28   \\
                    & \multicolumn{1}{c|}{FNN}  & 5.20  & 5.37  & 4.20  & 4.29  & 3.03   &                     & \multicolumn{1}{c|}{FNN}  & 5.38  & 5.82  & 5.57  & 4.77 & 4.28   \\
                    & \multicolumn{1}{c|}{SVR}  & 6.02  & 5.59  & 6.05  & 4.50  & 5.21  &                     & \multicolumn{1}{c|}{SVR}  & 6.27  & 6.35  & 6.45  & 5.42 & 5.39   \\
                    & \multicolumn{1}{c|}{ELM}  & 11.21 & 9.32 & 6.55  & 10.97 & 7.36   &                     & \multicolumn{1}{c|}{ELM}  & 10.43 & 6.81  & 8.31  & 9.42 & 8.54   \\ \hline
\multirow{7}{*}{S2} & \multicolumn{1}{c|}{Ours} & 7.01  & 4.47  & 3.71  & 4.95  & 2.56   & \multirow{7}{*}{S5} & \multicolumn{1}{c|}{Ours} & 5.41  & 5.23  & 3.98  & 3.87 & 4.41   \\
                    & \multicolumn{1}{c|}{LSTM} & 6.15  & 3.23  & 6.23  & 2.76  & 4.84   &                     & \multicolumn{1}{c|}{LSTM} & 2.75  & 3.59  & 2.97  & 2.25 & 2.97   \\
                    & \multicolumn{1}{c|}{GRU}  & 7.21  & 6.10  & 2.89  & 5.81  & 5.32   &                     & \multicolumn{1}{c|}{GRU}  & 2.61  & 4.26  & 2.83  & 2.71 & 4.02   \\
                    & \multicolumn{1}{c|}{CNN}  & 6.89  & 3.34  & 3.01  & 4.74 & 3.61   &                     & \multicolumn{1}{c|}{CNN}  & 3.09  & 4.34  & 2.79  & 2.79 & 4.36   \\
                    & \multicolumn{1}{c|}{FNN}  & 6.25  & 4.27  & 5.29  & 5.59  & 2.97   &                     & \multicolumn{1}{c|}{FNN}  & 4.57  & 4.61  & 4.53  & 4.01 & 3.82   \\
                    & \multicolumn{1}{c|}{SVR}  & 7.35  & 5.58  & 5.93  & 6.01  & 4.31   &                     & \multicolumn{1}{c|}{SVR}  & 8.65  & 5.34  & 6.16  & 5.79 & 5.75   \\
                    & \multicolumn{1}{c|}{ELM}  & 13.31      & 11.38 & 7.27  & 12.89 & 9.94   &                     & \multicolumn{1}{c|}{ELM}  & 9.12  & 10.33  & 7.24  & 8.29 & 11.74  \\ \hline
\multirow{7}{*}{S3} & \multicolumn{1}{c|}{Ours} & 5.43  & 3.74  & 4.91  & 3.91  & 3.20  & \multirow{7}{*}{S6} & \multicolumn{1}{c|}{Ours} & 3.21  & 3.27  & 5.61  & 6.28 & 3.97   \\
                    & \multicolumn{1}{c|}{LSTM} & 5.08  & 5.39  & 7.89  & 2.97  & 1.89  &                     & \multicolumn{1}{c|}{LSTM} & 2.79  & 2.71  & 3.65  & 2.35 & 1.23   \\
                    & \multicolumn{1}{c|}{GRU}  & 6.49  & 5.63  & 6.65  & 3.89  & 3.51   &                     & \multicolumn{1}{c|}{GRU}  & 2.86  & 5.80  & 5.83  & 3.26 & 1.83   \\
                    & \multicolumn{1}{c|}{CNN}  & 6.45  & 5.81  & 7.52  & 3.54  & 2.78   &                     & \multicolumn{1}{c|}{CNN}  & 2.83  & 9.22  & 6.52  & 4.69 & 3.45   \\
                    & \multicolumn{1}{c|}{FNN}  & 6.25  & 4.27  & 5.29  & 5.59  & 2.97   &                     & \multicolumn{1}{c|}{FNN}  & 3.97  & 5.11  & 5.50  & 6.24 & 4.32   \\
                    & \multicolumn{1}{c|}{SVR}  & 7.63  & 6.91  & 7.97  & 6.25  & 3.21   &                     & \multicolumn{1}{c|}{SVR}  & 7.09  & 6.19  & 7.15  & 7.72 & 2.97  \\
                    & \multicolumn{1}{c|}{ELM}  & 8.91  & 7.69  & 10.61 & 10.25 & 10.58  &                     & \multicolumn{1}{c|}{ELM}  & 7.66  & 12.24 & 10.13 & 6.91 & 7.31   \\ \hline
\end{tabular}
} 
\end{table*}
Fig. \ref{fig:2} demonstrates the evolution of the identified physiological parameters during the training of the proposed method. In Fig. \ref{fig:2}, the blue solid line illustrates the variation process of the parameters and the black dashed line indicates the estimated value by the proposed method which is the final convergent value of evolution. According to Table \ref{tab:table1} and Fig. \ref{fig:2}, the identified physiological parameters are within the physiologically acceptable range, indicating that the muscle forces calculated by the personalized Hill muscle model embedded in the proposed method are reasonable, which would directly benefit the guidance of the muscle force prediction.
\subsection{Overall Comparison}

\begin{figure*}
    \centering
    \includegraphics[width=\textwidth]{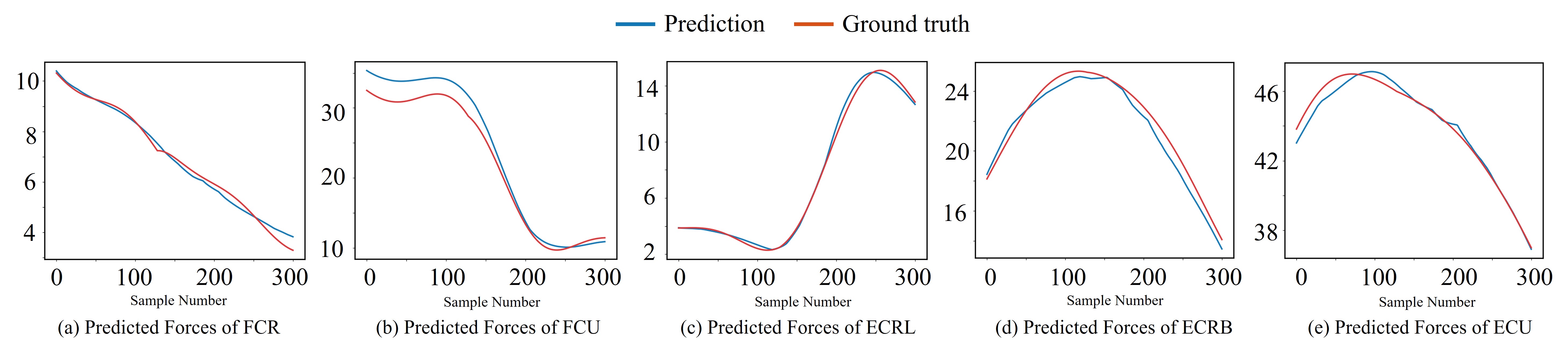}
    \caption{Representative results of the wrist case through the proposed method. The predicted outputs include FCR muscle force, FCU muscle force, ECRL muscle force, ECRB muscle force, and ECU muscle force.}
    \label{fig:3}
\end{figure*}

\begin{table*}[ht] 
\label{tab:overallr2}
\centering
\caption{R$^2$ of the proposed method and baseline methods of muscle forces prediction (wrist case)}
\resizebox{\textwidth}{!}{
\begin{tabular}{c|cccccc|c|cccccc}
\hline
Subjects             & Methods                   & FCR     & FCU     & ECRL    & ECRB    & ECU    & Subjects             & Methods                   & FCR     & FCU     & ECRL    & ECRB   & ECU     \\ \hline
\multirow{7}{*}{S1} & \multicolumn{1}{c|}{Ours} & 0.98  & 0.97  & 0.98  & 0.96  & 0.96   & \multirow{7}{*}{S4} & \multicolumn{1}{c|}{Ours} & 0.97  & 0.97  & 0.97  & 0.97 & 0.97   \\
                    & \multicolumn{1}{c|}{LSTM} & 0.98  & 0.99  & 0.98  & 0.98  & 0.96   &                     & \multicolumn{1}{c|}{LSTM} & 0.98  & 0.98  & 0.97  & 0.96 & 0.95   \\
                    & \multicolumn{1}{c|}{GRU}  & 0.98  & 0.98  & 0.99  & 0.97  & 0.95   &                     & \multicolumn{1}{c|}{GRU}  & 0.97  & 0.98  & 0.96  & 0.96 & 0.96   \\
                    & \multicolumn{1}{c|}{CNN}  & 0.98  & 0.98  & 0.99  & 0.97  & 0.96   &                     & \multicolumn{1}{c|}{CNN}  & 0.96  & 0.97  & 0.97  & 0.96 & 0.94   \\
                    & \multicolumn{1}{c|}{FNN}  & 0.98  & 0.96  & 0.98  & 0.98  & 0.96   &                     & \multicolumn{1}{c|}{FNN}  & 0.97  & 0.96  & 0.97  & 0.96 & 0.96   \\
                    & \multicolumn{1}{c|}{SVR}  & 0.97  & 0.96  & 0.96  & 0.98  & 0.94   &                     & \multicolumn{1}{c|}{SVR}  & 0.96  & 0.96  & 0.96  & 0.96 & 0.95   \\
                    & \multicolumn{1}{c|}{ELM}  & 0.92  & 0.89  & 0.96  & 10.90 & 0.92   &                     & \multicolumn{1}{c|}{ELM}  & 0.91 & 0.96  & 0.94  & 0.91 & 0.92   \\ \hline
\multirow{7}{*}{S2} & \multicolumn{1}{c|}{Ours} & 0.96  & 0.97  & 0.98  & 0.97  & 0.97  & \multirow{7}{*}{S5} & \multicolumn{1}{c|}{Ours} & 0.96  & 0.97  & 0.98  & 0.97 & 0.95   \\
                    & \multicolumn{1}{c|}{LSTM} & 0.97  & 0.98  & 0.96  & 0.99  & 0.96   &                     & \multicolumn{1}{c|}{LSTM} & 0.99  & 0.99  & 0.99  & 0.99 & 0.98  \\
                    & \multicolumn{1}{c|}{GRU}  & 0.96  & 0.96  & 0.99  & 0.96  & 0.95   &                     & \multicolumn{1}{c|}{GRU}  & 0.99  & 0.99  & 0.99  & 0.99 & 0.96   \\
                    & \multicolumn{1}{c|}{CNN}  & 0.96  & 0.98  & 0.99  & 0.97  & 0.97   &                     & \multicolumn{1}{c|}{CNN}  & 0.99  & 0.99  & 0.99  & 0.99 & 0.96   \\
                    & \multicolumn{1}{c|}{FNN}  & 0.97  & 0.97  & 0.97  & 0.96  & 0.98   &                     & \multicolumn{1}{c|}{FNN}  & 0.97  & 0.97  & 0.97  & 0.96 & 0.98   \\
                    & \multicolumn{1}{c|}{SVR}  & 0.95  & 0.96  & 0.96  & 0.96  & 0.96   &                     & \multicolumn{1}{c|}{SVR}  & 0.95  & 0.97  & 0.96  & 0.96 & 0.95   \\
                    & \multicolumn{1}{c|}{ELM}  & 0.91  & 0.91  & 0.94  & 0.88  & 0.90   &                     & \multicolumn{1}{c|}{ELM}  & 0.95  & 0.93  & 0.95  & 0.94 & 0.90  \\ \hline
\multirow{7}{*}{S3} & \multicolumn{1}{c|}{Ours} & 0.98  & 0.98  & 0.97  & 0.98  & 0.96   & \multirow{7}{*}{S6} & \multicolumn{1}{c|}{Ours} & 0.99  & 0.98  & 0.97  & 0.94 & 0.95   \\
                    & \multicolumn{1}{c|}{LSTM} & 0.98  & 0.97  & 0.95  & 0.98  & 0.99   &                     & \multicolumn{1}{c|}{LSTM} & 0.99  & 0.99  & 0.99  & 0.99 & 0.99   \\
                    & \multicolumn{1}{c|}{GRU}  & 0.97  & 0.97  & 0.96  & 0.97  & 0.94   &                     & \multicolumn{1}{c|}{GRU}  & 0.99  & 0.96  & 0.97  & 0.98 & 0.99   \\
                    & \multicolumn{1}{c|}{CNN}  & 0.97  & 0.97  & 0.95  & 0.98  & 0.96   &                     & \multicolumn{1}{c|}{CNN}  & 0.99  & 0.93  & 0.96  & 0.96 & 0.95   \\
                    & \multicolumn{1}{c|}{FNN}  & 0.97  & 0.98  & 0.97  & 0.96  & 0.96   &                     & \multicolumn{1}{c|}{FNN}  & 0.98  & 0.97  & 0.97  & 0.94 & 0.94   \\
                    & \multicolumn{1}{c|}{SVR}  & 0.95  & 0.96  & 0.95  & 0.96  & 0.94   &                     & \multicolumn{1}{c|}{SVR}  & 0.94  & 0.96  & 0.95  & 0.93 & 0.98   \\
                    & \multicolumn{1}{c|}{ELM}  & 0.94  & 0.94  & 0.92 & 0.91 & 0.86  &                     & \multicolumn{1}{c|}{ELM}  & 0.94  & 0.90 & 0.92 & 0.94 & 0.91   \\ \hline
\end{tabular}
} 
\end{table*}

\begin{figure}
\centering
\begin{adjustbox}{center}
  \includegraphics[scale=0.36]{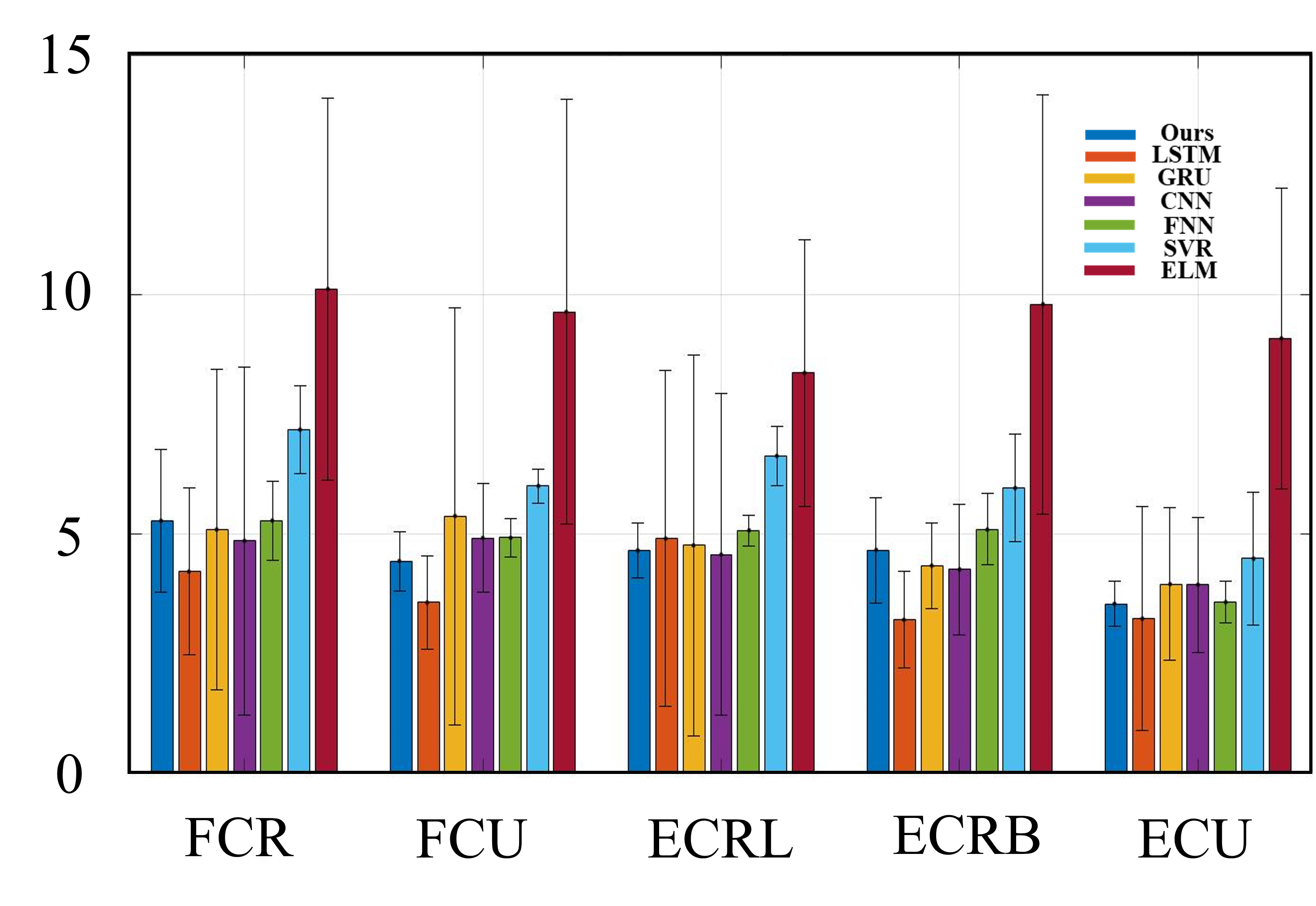}
\end{adjustbox}
\caption{Average RMSEs of the included muscle forces across all the subjects (wrist case).}
\label{fig:5}
\end{figure}

\begin{figure*}
  \centering
  \scriptsize
  \includegraphics[scale=0.4]{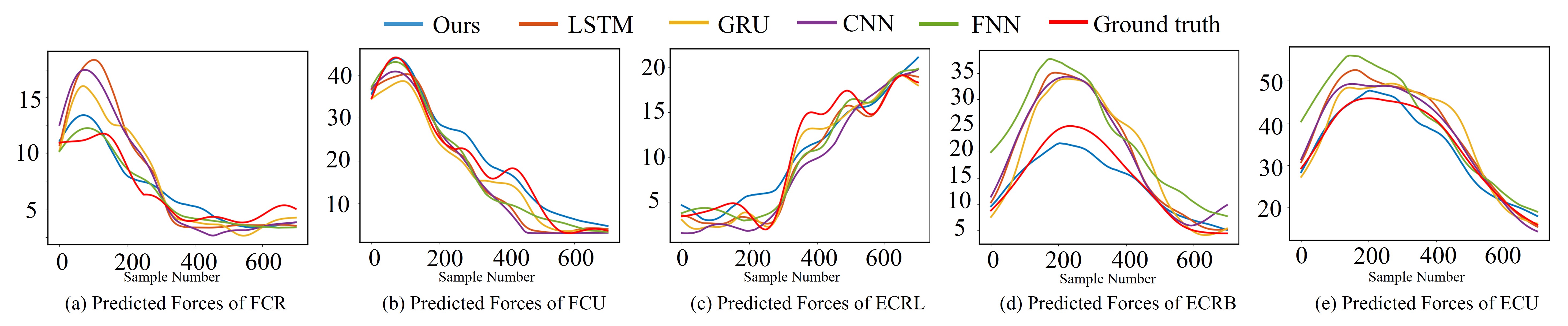} 
  \caption{Comparison results of the proposed method and baseline methods in the intrasession scenario.}
  \label{fig:4}  
\end{figure*}

For the prediction of muscle forces, the proposed method uses the unlabeled sEMG data in the training phase, while the baseline methods use the labeled sEMG data. Fig. \ref{fig:3} shows the representative results of the proposed method for the prediction of muscle forces FCR, FCU, ECRL, ECRB, and ECU. According to Fig. \ref{fig:3}, we can find the proposed method could predict the muscle forces well. 

Detailed comparisons of all the subjects between the proposed method and baseline methods are presented in Table III and Table IV. In the experiment, we use the data with the same flexion speed to train and test the proposed method and baseline methods. We randomly select 70\% of the data for training, while the rest 30\% for testing. The number of training data is 10500, and the number of testing data is 4500. According to Table III and Table IV, deep learning-based methods, including the proposed method, LSTM, GRU, CNN and FNN, achieve better-predicted performance than machine learning-based methods, i.e., SVR and ELM, as evidenced by smaller RMSEs and higher $R^2$ in most cases. Because these deep learning-based methods could automatically extract high-level features from the collected data. Furthermore, the proposed method could achieve comparable performance with LSTM and GRU in some situations with unlabeled data, and the performance of the proposed method is better than that of FNN, which indicates the effectiveness of the designed loss function.

Fig. \ref{fig:5} shows the average RMSEs of muscle forces prediction of the proposed method and baseline methods. The proposed method achieves an overall performance similar to that of LSTM, GRU, CNN and FNN without direct reliance on actual muscle force labels. 
In the training process, FNN used in the proposed method is not only trained based on the MSE loss but also enhanced by the physics-informed losses. The embedded physics laws provide the potential relationships between the output variables as learnable features for the training. 

Table \ref{table:12} details the training time of deep learning-based methods, including GRU, LSTM, CNN, FNN and the proposed method. Accordingly, for all the methods, the training time is less with the increase of the batch size, and the proposed method has the longest training time compared to other baseline methods. This is because the proposed method is developed under the PINN framework, it not only involves the minimization of the MSE of FNN but also the regularization of physics-derived terms.  
\begin{table}[ht]\small%
\centering
\scriptsize
\caption{Training time of deep learning-based methods (h)}
\resizebox{0.5\textwidth}{!}{
\begin{tabular}{ccccccc}
\toprule
\multicolumn{2}{c}{Methods}    & GRU  & LSTM & CNN  & FNN  & Ours  \\ \midrule
\multirow{3}{*}{Batch Sizes} & 1 & 5.53 & 5.69 & 3.97 & 4.15 & 10.52 \\
                            & 4 & 3.81 & 3.99 & 3.11 & 3.67 & 9.21 \\
                            & 8 & 3.67 & 3.74 & 2.87 & 3.25 & 8.35  \\ \bottomrule
\end{tabular}
}
\label{table:12}
\end{table}

\subsection{Evaluation of Intrasession Scenario}
The performance of the proposed method in the intrasession scenario is also demonstrated to validate its robustness. 
For each subject, we train the proposed method and baseline methods with the data of one flexion speed and then test them using the data of another flexion speed. In the experiment, we only demonstrate the comparison results of the proposed method and deep learning methods in Fig. \ref{fig:4} to make these results clearer.
According to Fig. \ref{fig:4}, the proposed method demonstrates exceptional performance in datasets with different distributions, but the predicted results of some baseline methods are degraded. In particular, concerning the predicted results of muscle forces of ECRL and ECU, the predicted results yielded by the proposed method demonstrate a notably enhanced congruence with the underlying ground truth.
LSTM, GRU, and CNN demonstrate the ability of motion pattern recognition since their muscle force prediction curves are generally consistent with the trend of ground truth.
Additionally, these methods exhibit the proficiency of dynamical tracking in part of the predicted results but the error remains in other predicted results, especially when it comes to capturing peak and trough values, noticeable discrepancies can be observed in the predicted values, which reflects the limitation of the stability.
Specifically, it demonstrates strong performance in the prediction of FCR, FCU, and ECRL, while it still exhibits significant discrepancies in the prediction of the ECRB and ECU.
The proposed method manifests a discernible capability to predict muscle forces on data characterized by the diverse distribution without label information.

\subsection{Effects of Network Architectures}
To investigate the effects of network architectures on performance, we implement the proposed method with different numbers of FC blocks. Table \ref{table:2} lists the detailed comparison results, we can find the proposed method could achieve the best performance with four FC blocks. Although the increase in the number of FC blocks would help extract more representative features, the proposed method may be overfitting when we continue to add FC blocks, which degrades its performance.

\begin{table}[ht]\small%
\centering
\scriptsize
\caption{Comparisons of the proposed method with different number of FC blocks (R$^2$)}
\resizebox{0.5\textwidth}{!}{
\begin{tabular}{@{}c*{5}{c}@{}}
\toprule
Number of FC Blocks & FCR & FCU & ECRL & ECRB & ECU  \\ \midrule
3                   & 0.96   & 0.82  & 0.94    & 0.97    & 0.87         \\
4                   & 0.98   & 0.97  & 0.96    & 0.97    & 0.97        \\
5                   & 0.98   & 0.87  & 0.96    & 0.97    & 0.90         \\
6                   & 0.96   & 0.90  & 0.95    & 0.96    & 0.94         \\ \bottomrule
\end{tabular}
}
\label{table:2}
\end{table}

\begin{table}[h]\tiny%
\centering
\caption{Comparisons of the proposed method with different learning rates (R$^2$)}
\resizebox{0.5\textwidth}{!}{
\renewcommand{\arraystretch}{0.8}
\begin{tabular}{@{}c*{5}{c}@{}}
\toprule
Learning Rates & FCR & FCU & ECRL & ECRB & ECU \\ \midrule
0.01           & 0.96   & 0.93  & 0.95    & 0.94    & 0.93        \\
0.001          & 0.98   & 0.97  & 0.96    & 0.97    & 0.97       \\
0.0001         & 0.98   & 0.96  & 0.94    & 0.97    & 0.97        \\ 
\bottomrule
\end{tabular}
}
\label{table:3}
\end{table}

\begin{table}
\centering
\scriptsize
\caption{Comparisons of the proposed method with different activation functions (R$^2$)}
\resizebox{0.5\textwidth}{!}{
\begin{tabular}{@{}c*{5}{c}@{}}
\toprule
Activation Functions & FCR & FCU & ECRL & ECRB & ECU \\ \midrule
Sigmoid & 0.16 & 0.21 & 0.15 & 0.17 & 0.09 \\
Tanh & 0.26 & 0.20 & 0.37 & 0.17 & 0.30 \\
ReLU & 0.98 & 0.97 & 0.96 & 0.97 & 0.97 \\
Leaky ReLU & 0.95 & 0.94 & 0.93 & 0.92 & 0.91 \\
ELU & 0.89 & 0.88 & 0.87 & 0.86 & 0.85 \\
\bottomrule
\end{tabular}
}
\label{table:4}
\end{table}

\begin{table}[ht]\tiny%
\centering
\caption{Comparisons of the proposed method with different batch sizes (R$^2$)}
\resizebox{0.5\textwidth}{!}{
\renewcommand{\arraystretch}{0.7}
\begin{tabular}{@{}c*{5}{c}@{}}
\toprule
Batch Sizes & FCR & FCU & ECRL & ECRB & ECU \\ \midrule
1 & 0.98 & 0.97 & 0.96 & 0.97 & 0.97 \\
8 & 0.90 & 0.92 & 0.91 & 0.89 & 0.90 \\
16 & 0.83 & 0.97 & 0.96 & 0.90 & 0.97 \\
32 & 0.82 & 0.96 & 0.96 & 0.91 & 0.97 \\  
64 & 0.79 & 0.88 & 0.87 & 0.86 & 0.88 \\ \bottomrule
\end{tabular}
}
\label{table:5}
\end{table}

\subsection{Effects of Network Parameters}
We also consider the effects of network parameters on the performance, including batch size, learning rate and type of activation function. Table \ref{table:3} shows R$^2$ of the proposed method with different learning rates, it seems that its R$^2$ is without obvious fluctuations. Table \ref{table:4} lists R$^2$ of the proposed method with different types of activation functions, we can find when ReLU is selected as the activation function, the proposed method has the best performance. Table \ref{table:5} shows the effects of different batch sizes. When the batch size is 1, the proposed method achieves better performance, because the network could learn the representations better.

\subsection{Effects of Training Data Number}
Table \ref{table:number} shows the experimental results of the proposed method with a different number of training data. When the number of training data is more than 10500, the proposed method achieves satisfactory performance with little fluctuations.
Increasing the training data beyond 10500 samples does not significantly enhance the performance of the proposed method, as evidenced by the minimal improvements seen with 14500 and 17000 samples. Such findings highlight the importance of a balanced approach to data collection and model training, emphasizing data quality and representativeness over sheer quantity. 

\begin{table}[ht]\scriptsize%
\centering
\caption{Comparisons of the proposed method with different training data number (R$^2$)}
\resizebox{0.5\textwidth}{!}{
\begin{tabular}{@{}c*{5}{c}@{}}
\toprule
Training data numbers  & FCR & FCU & ECRL & ECRB & ECU  \\ \midrule
3500                     & 0.81   & 0.79  & 0.74    & 0.82    & 0.90         \\
7000                    & 0.89   & 0.91  & 0.90    & 0.87    & 0.89        \\
10500                  & 0.97   & 0.94  & 0.95    & 0.96    & 0.94         \\
14000                  & 0.97   & 0.98  & 0.96    & 0.98    & 0.95         \\ 
17500                 & 0.98   & 0.97  & 0.97    & 0.96    & 0.97         \\
\bottomrule
\end{tabular}
}
\label{table:number}
\end{table}

\section{Discussion}\label{discussions}
\begin{table}[ht]
\normalsize
\caption{Identified physiological parameters of the specific subject (knee case)}
\centering
\renewcommand{\arraystretch}{1} 
\resizebox{0.5\textwidth}{!}{
\begin{tabular}{cccc}
\toprule
\multicolumn{1}{c}{Parameter Indices} & Muscle Indices & Estimations & Physiological Ranges \\ \midrule
\multirow{2}{*}{$l^{m}_{o}$(m)}                & BFS          & 0.1819     & 0.1630 - 0.1830     \\
                                    & RF           & 0.1143     & 0.1040 - 0.1240     \\ \hline
\multirow{2}{*}{$F^{m}_{o}$(N)}                & BFS          & 805.7      & 402 - 1206          \\
                                    & RF           & 1199.6     & 584.5 - 1753.5      \\ \bottomrule
\end{tabular}
}
\label{tab:kneepar}
\end{table}

\begin{figure}
    \centering 
    \includegraphics[width=0.48\textwidth]{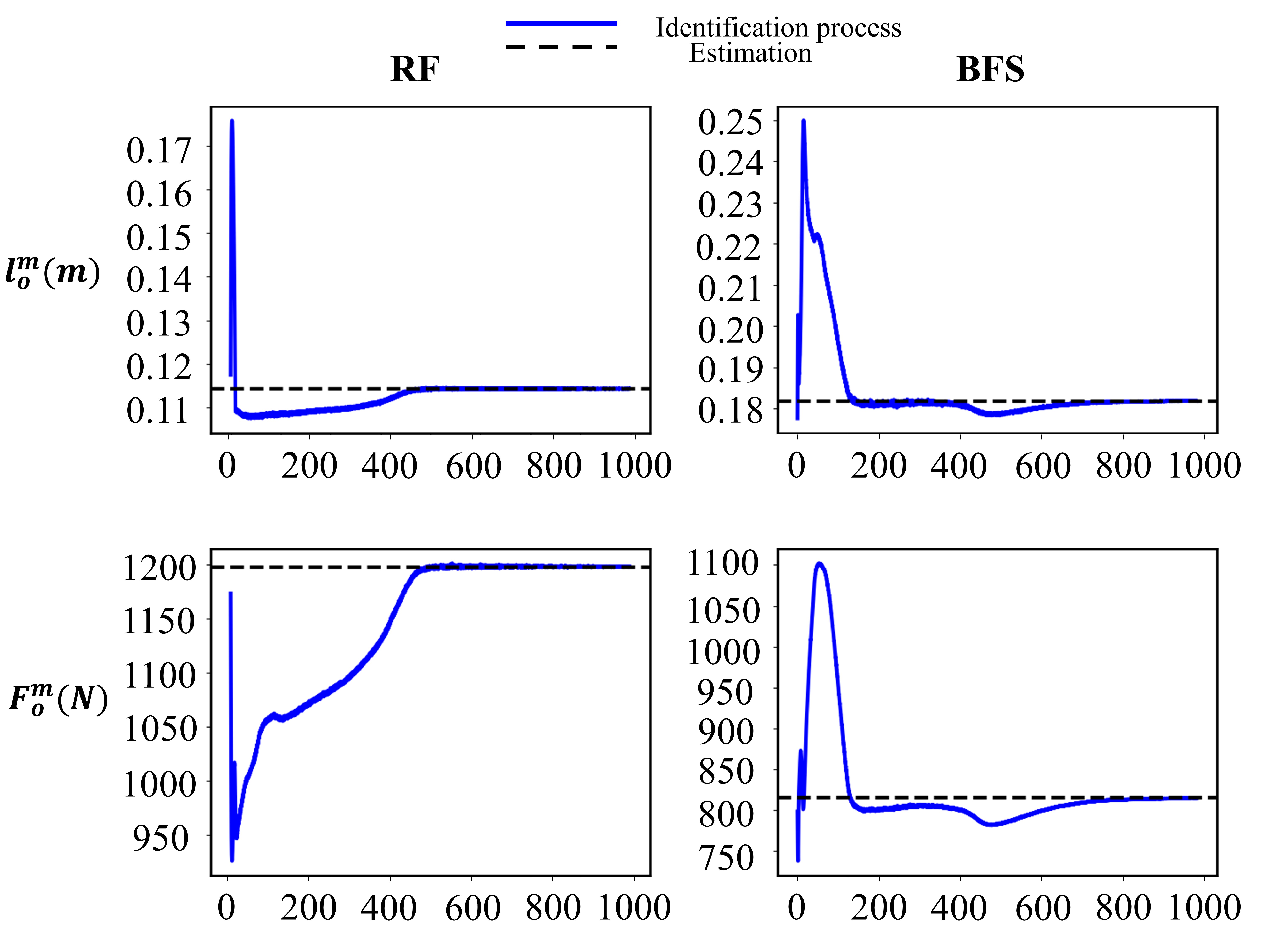}
    \caption{Evolution of the maximum isometric muscle force $F_0^m$ and the optimal muscle fiber length $l_0^m$ identified of the specific subject during the training of the proposed method in the knee case (The blue solid line illustrates the variation process of the parameters and the black dashed line indicates the estimated value by the proposed method which is the final convergent value of evolution).}
    \label{fig:kneepar}
\end{figure}

In this section, we discuss the generalization of the proposed method, and potential ways to further enhance its performance from various aspects.

\begin{figure} 
    \centering 
    \includegraphics[width=0.47\textwidth]{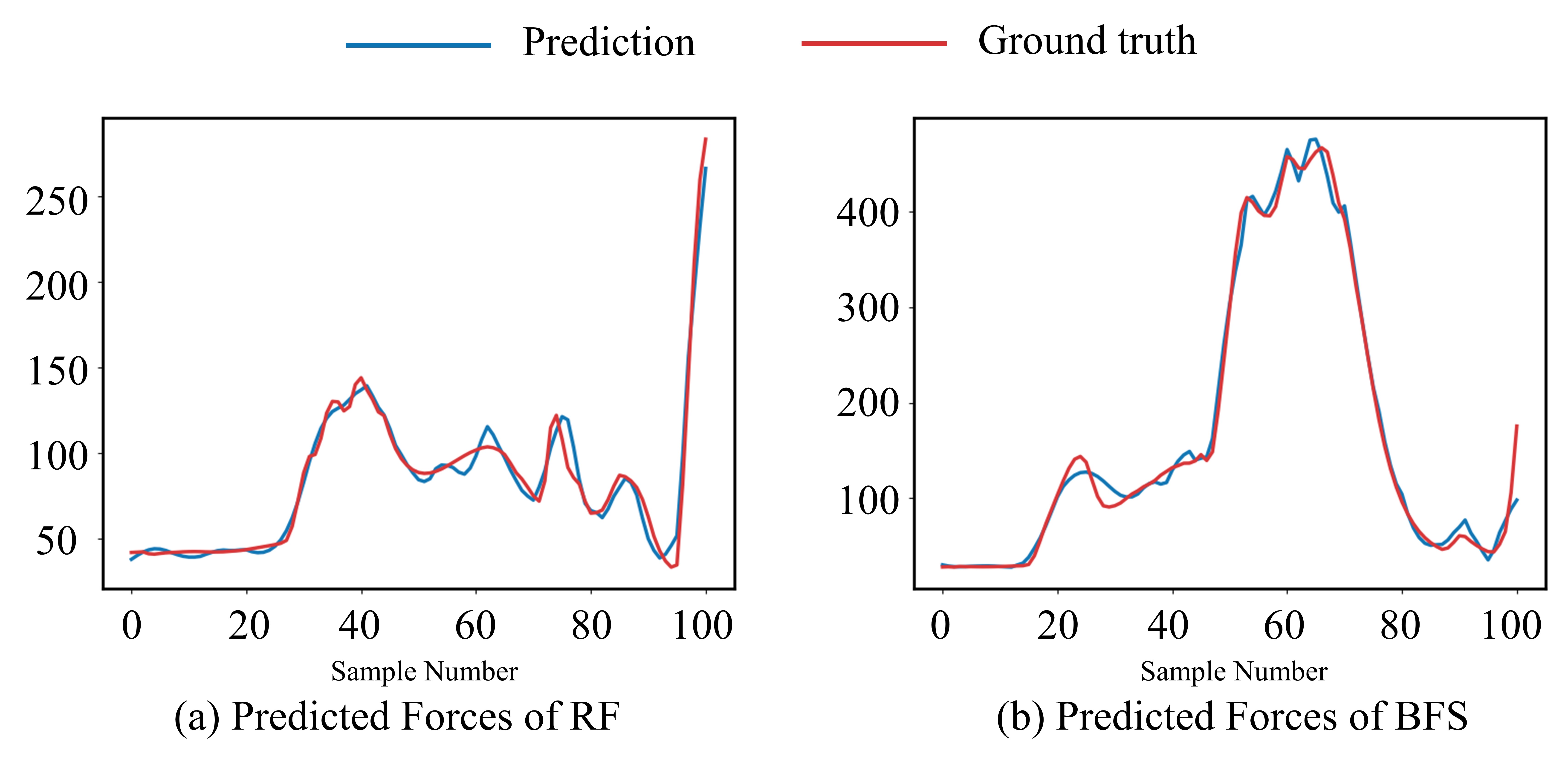}
    \caption{Representative results of the knee case of the proposed method. The predicted outputs include BFS muscle force and RF muscle force.}
    \label{fig:kneeforce}
\end{figure}

In this paper, we only use muscle forces prediction of wrist flexion/extension as an example to demonstrate the feasibility and effectiveness of the proposed method. Actually, the proposed method can also be generalized to other joints. Table \ref{tab:kneepar}, Table \ref{tab:kneeoverall}, Fig. \ref{fig:kneepar} and Fig. \ref{fig:kneeforce} show the details of physiological parameter identification and muscle forces prediction (including biceps femoris short head (BFS) and rectus femoris (RF)) of knee flexion/extension. To be specific, Table \ref{tab:kneepar} and Fig. \ref{fig:kneepar} show the results of the identified physiological parameters, we can find all these physiological parameters are within the physiologically acceptable range. Additionally, Table \ref{tab:kneeoverall} and Fig. \ref{fig:kneeforce} detail the predicted results of BFS and RF. Accordingly, the proposed method can fit the ground truth curve well and obtain comparable predictions compared with FNN even without any label information.
\begin{table}
\tiny
\centering
\caption{RMSE and $R^2$ of the proposed method and baseline methods of muscle forces prediction (Knee case)}
\resizebox{0.47\textwidth}{!}{%
\begin{tabular}{cc|cc}
\hline
Subjects & Methods & RF & BFS \\ \hline
\multirow{2}{*}{S1} & Ours    & 9.33/0.96   & 15.53/0.98    \\
                    & FNN     & 12.29/0.93  & 16.18/0.98    \\ \hline
\multirow{2}{*}{S2} & Ours    & 20.27/0.93  & 26.96/0.91    \\
                    & FNN     & 18.41/0.94  & 18.06/0.93    \\ \hline
\multirow{2}{*}{S3} & Ours    & 16.71/0.93  & 22.43/0.90    \\
                    & FNN     & 13.25/0.94  & 28.19/0.89    \\ \hline
\multirow{2}{*}{S4} & Ours    & 32.47/0.93  & 21.29/0.97    \\
                    & FNN     & 28.62/0.94  & 23.71/0.96    \\ \hline
\multirow{2}{*}{S5} & Ours    & 17.27/0.95  & 15.45/0.94    \\
                    & FNN     & 22.51/0.94  & 19.85/0.92    \\ \hline
\multirow{2}{*}{S6} & Ours    & 17.39/0.94  & 14.12/0.94    \\
                    & FNN     & 24.32/0.93  & 14.97/0.95    \\ \hline
\end{tabular}%
}
\label{tab:kneeoverall}
\end{table}

During the implementation of the proposed method, we partially simplify the MSK forward dynamics model by reducing the number of individualized physiological parameters. Only the maximum isometric muscle force and the optimal fiber length are considered to be identified, and all the other physiological parameters are directly derived from the scaled wrist model. Moreover, five primary muscles have been selected as the key actuators for wrist flexion/extension, but these muscle-tendon units may also affect other degrees of freedom in wrist movements. In the future, we will try to relax these simplifications and assumptions by considering more physiological parameters and physics laws to obtain a more physiologically accurate representation of muscle tissues with connective tissues and muscle fibers, making it more feasible in practical and clinical applications. The computational time of the proposed method is longer than baseline methods because it is developed under the physics-informed neural network framework, which not only involves the minimization of the MSE of FNN but also the regularization of physics-informed terms during the network training. In the future, we will design a distributed framework for the proposed method to accelerate its training, and also consider pre-training an initial model with subject-specific data and then updating the model with other subjects' data, which can simultaneously reduce the training time and enhance the generalization.

\section{Conclusion}\label{conclusion}
This paper presents a novel physics-informed deep-learning method mainly for muscle forces estimation with unlabeled sEMG data, and the proposed method could simultaneously identify parameters of the Hill muscle model. Specifically, the proposed method uses the MSK forward dynamics as the residual loss for the identification of personalized physiological parameters and another residual constraint based on the muscle contraction dynamics for the estimation of muscle forces without data labels. Comprehensive experiments indicate the feasibility of the proposed method. 
\bibliographystyle{ieeetr}
\bibliography{reference}

\vspace{12pt}
\color{red}

\end{document}